\documentclass[conference]{IEEEtran}
\IEEEoverridecommandlockouts

\usepackage{cite}
\usepackage{amsmath,amssymb,amsfonts}
\usepackage{algorithmic}
\usepackage{graphicx}
\usepackage{textcomp}
\usepackage{xcolor}
\usepackage{mathtools}
\usepackage{orcidlink}
\usepackage{verbatim}
\def\BibTeX{{\rm B\kern-.05em{\sc i\kern-.025em b}\kern-.08em
    T\kern-.1667em\lower.7ex\hbox{E}\kern-.125emX}}
\begin{document}

\IEEEpubid{\begin{minipage}{\textwidth}\centering
    \vspace{2.0cm} 
    \footnotesize
    \copyright~2025 IEEE. Personal use of this material is permitted.  Permission from IEEE must be obtained for all other uses, in any current or future media, including reprinting/republishing this material for advertising or promotional purposes, creating new collective works, for resale or redistribution to servers or lists, or reuse of any copyrighted component of this work in other works.
\end{minipage}}

\title{Interpreting Deep Neural Network-Based Receiver Under Varying Signal-To-Noise Ratios}

\author{
    \IEEEauthorblockN{1\textsuperscript{st} Marko Tuononen
    \orcidlink{0009-0008-3069-1777}}
    \IEEEauthorblockA{\textit{Nokia Networks} \\
    \textit{Nokia Group}\\
    Espoo, Finland }
    \and
    \IEEEauthorblockN{2\textsuperscript{nd} Dani Korpi
    \orcidlink{0000-0003-3460-7436}}
    \IEEEauthorblockA{\textit{Nokia Bell Labs} \\
    \textit{Nokia Group}\\
    Espoo, Finland }
    \and
    \IEEEauthorblockN{3\textsuperscript{rd} Ville Hautamäki
    \orcidlink{0000-0002-5885-0003}}
    \IEEEauthorblockA{\textit{School of Computing} \\
    \textit{University of Eastern Finland}\\
    Joensuu, Finland }
    \thanks{Ville Hautamäki was partially supported by Jane and Aatos Erkko Foundation. The authors would also like to thank Jyri Suvanen for his contributions.}
}
\maketitle

\begin{abstract}
We propose a novel method for interpreting neural networks, focusing on convolutional neural network-based receiver model. The method identifies which unit or units of the model contain most (or least) information about the channel parameter(s) of the interest, providing insights at both global and local levels---with global explanations aggregating local ones. Experiments on link-level simulations demonstrate the method's effectiveness in identifying units that contribute most (and least) to signal-to-noise ratio processing. Although we focus on a radio receiver model, the method generalizes to other neural network architectures and applications, offering robust estimation even in high-dimensional settings.
\end{abstract}

\begin{IEEEkeywords}
Interpretable Machine Learning, Neural Network Interpretation, Convolutional Neural Networks, Radio Receiver
\end{IEEEkeywords}

\section{Introduction}
\label{sec:intro}
Neural networks are often too complex for direct human interpretation, as a single prediction can involve billions of mathematical operations and weights. Therefore, specific methods have been developed to interpret neural networks, understand their learning processes, extract additional information, justify their decisions, and evaluate these aspects in the context of real-world problems \cite{molnar2022}.

Interpretability of neural network models is crucial for developers to troubleshoot and improve the models \cite{huyen2022}. Understanding how the model arrives at a particular decision helps identify and fix problems, enhancing overall performance. Interpretability is also important for users to build trust and detect potential biases \cite{huyen2022}. Knowing how a model makes decisions allows users to confidently rely on its outputs, fostering transparency and accountability. Additionally, regulations like the European Union Artificial Intelligence Act \cite{euaiact2024} and the AI Ethics Guidelines by the European Commission \cite{euaiethics2019} emphasize the importance of interpretability in AI systems.

\begin{figure}[ht]
\centering
\includegraphics[width=\columnwidth]{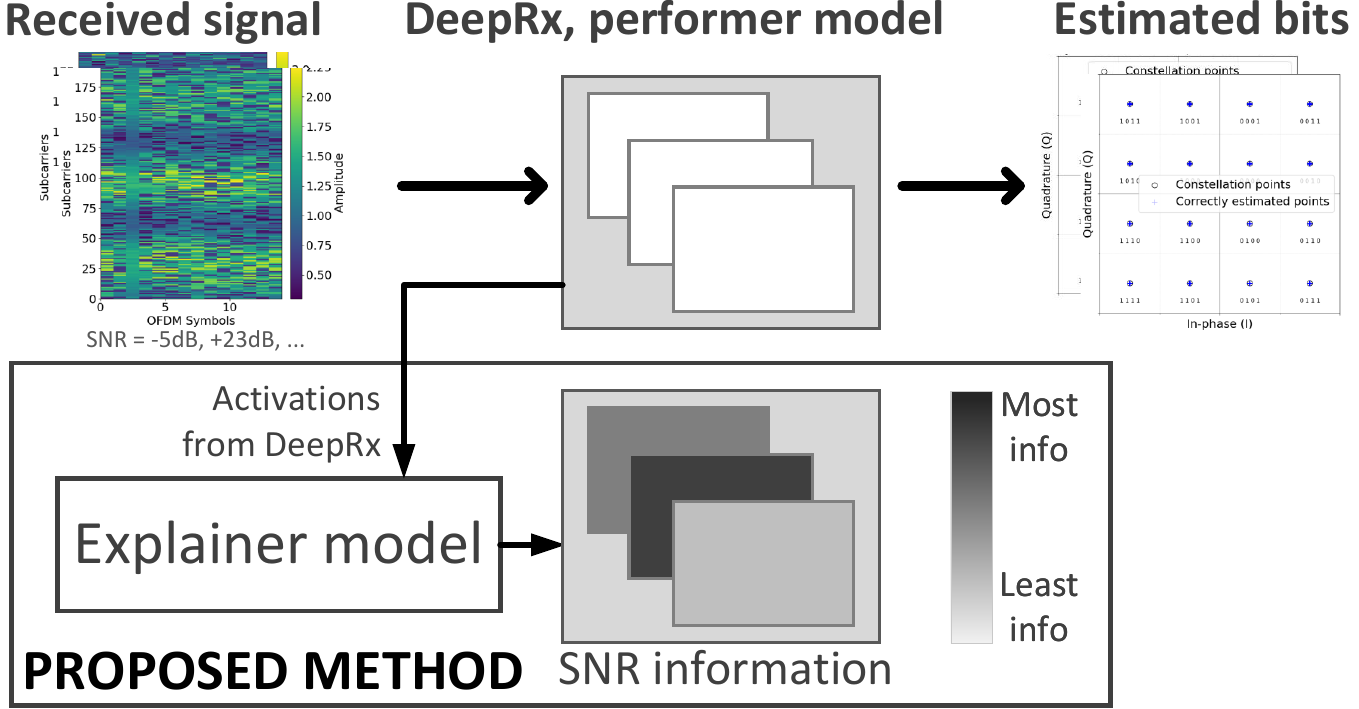}
\caption{Proposed method revealing how specific parts of the DeepRx model contain information about Signal-to-Noise Ratio processing.}
\label{fig:high_level}
\end{figure}

In wireless communication systems, channel parameters like Signal-to-Noise Ratio (SNR), Doppler spread, and delay spread describe the wireless channel's characteristics and quality. These metrics are essential for designing and evaluating the physical layer, affecting the system's quality, reliability, and performance \cite{tse2008, goldsmith2005}. Understanding and mitigating their impacts can lead to more robust and efficient systems.

Machine learning models are expected to gradually replace the classical signal processing in the physical layer \cite{hoydis2021}. Unlike traditional methods, these models do not explicitly model channel parameters. This paper focuses on one such proposed model-–-a deep neural network-based receiver model, known as DeepRx \cite{honkala2021deeprx}. DeepRx substitutes multiple signal processing blocks (channel estimation, equalization, and soft demapping) in the physical layer with a fully convolutional neural network. Trained on received waveforms in the frequency domain with corresponding transmitted bits as labels, DeepRx detects received bits and estimates their uncertainty across different modulation orders, ensuring 5G NR compliance.

While the varying behavior of DeepRx under different channel conditions is observable, its internal mechanisms to accommodate these diverse scenarios remain unclear. Identifying these mechanisms is crucial for improving, troubleshooting, and trusting the model in real-world applications. To this end, our main contribution is providing insight into the internal mechanisms of the deep neural network-based receiver model under varying SNRs. We propose a novel method to interpret and gain insights into model’s internal workings under varying SNRs, enhancing our understanding of its behavior.

\begin{figure*}[t]
\centering
\includegraphics[width=\textwidth]{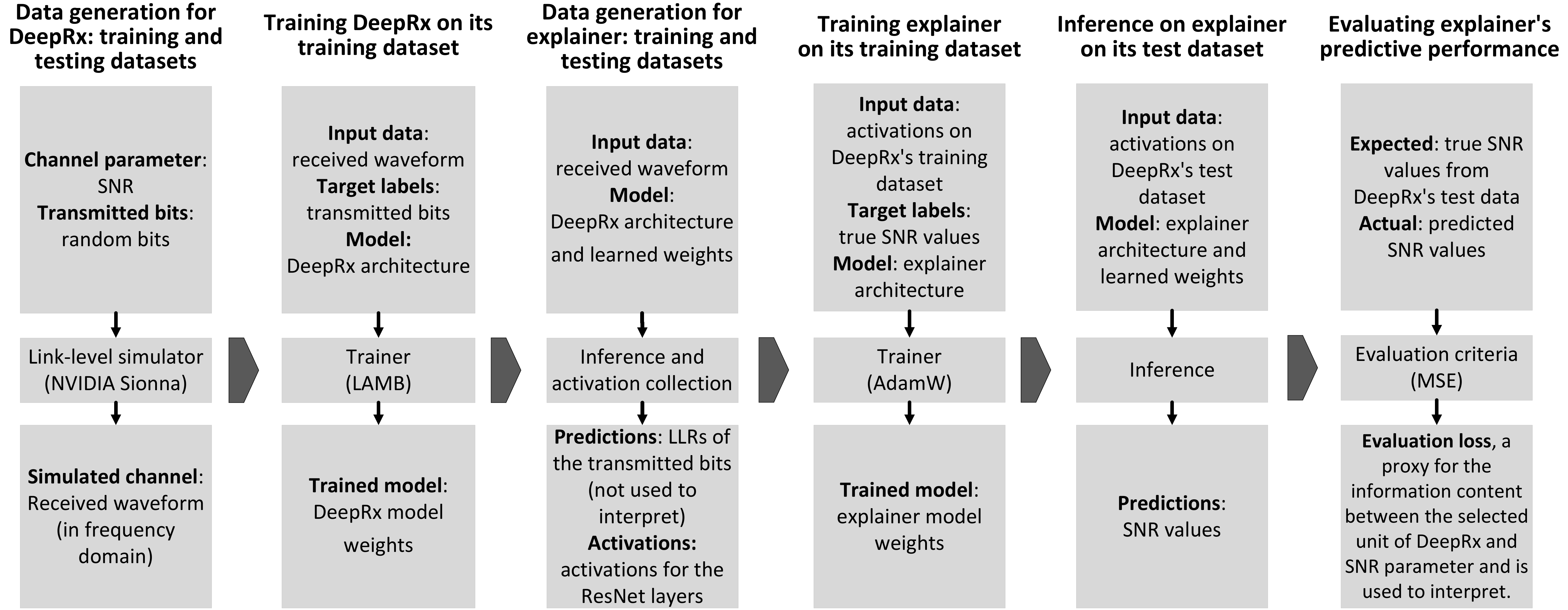}
\caption{Interpreting the internal processing of the deep neural network-based receiver model (DeepRx) under varying Signal-to-Noise Ratios.}
\label{fig:proposed_method}
\end{figure*}

\section{Related Work}
\label{sec:relatedwork}
Detecting how the deep neural network-based receiver model handles channel parameters is analogous to discovering abstract features and concepts in neural networks. Existing methods can be categorized into learned features \cite{olah2017, bau2017}, which refer to the high-level features learned in hidden layers; detecting concepts \cite{kim2018, ghorbani2019, koh2020, chen2020, zhang2018, bauerle2022}, which generate explanations that are not limited by the feature space; and other relevant methods \cite{zharov2020}. While these methods provide valuable insights, each suffers from one or more drawbacks. Our proposed method addresses these drawbacks as follows:

\begin{enumerate}
  \item \textbf{Domain Suitability}: Most methods are designed for image data and require modifications for other data types, making their application to the deep neural network-based receiver model unclear. Our method is suitable for the deep neural network-based receiver model without modifications.
  \item \textbf{Resource Efficiency}: Augmenting models to improve interpretability can deteriorate discrimination power and increase resource consumption during inference. Resource efficiency is critical for the deep neural network-based receiver model in practical applications. Our method maintains the discrimination power and resource consumption of the model.
  \item \textbf{Human-Relevant Interpretations}: Existing methods may not leverage the channel parameters essential for human experts in designing and evaluating wireless communication systems. Our method leverages the key channel parameters critical for human expert understanding and bases its interpretations on these parameters.
  \item \textbf{Complexity Capture}: Existing methods may not fully capture the complexity hidden in the deep neural network-based receiver model. Our method, due to the Universal Approximation Theorem \cite{sonoda2017}, has the potential to capture all the complexities hidden in the model.
\end{enumerate}

The proposed method can be viewed as an estimator for the Mutual Information (MI) between a unit or units of the deep neural network-based receiver model and selected channel parameter(s). Although the proposed method does not directly estimate MI, it uses predictive performance as a proxy for the MI between model units and channel parameters. Calculating exact MI between continuous random variables is generally infeasible due to the difficulty in obtaining the true joint probability distribution \cite{murphy2022}. Practical MI estimators include discretizing continuous variables \cite{scott1979}, using (K-)Nearest Neighbours to estimate densities \cite{kraskov2004}, and employing variational inference to approximate the conditional distribution and optimize a lower bound on MI \cite{poole2019}.

\section{Methodology}
\label{sec:methodology}
We propose a method for extracting interpretations from inside a deep neural network-based receiver model, \textit{a performer model}, by training another neural network, \textit{an explainer model}, in a supervised manner to predict channel parameter(s) from the activations of the explainer model and by evaluating the explainer model’s predictive performance.

The input data for the explainer model are the activations from a unit or units of the performer model. Unit can be individual neurons, convolutional channels, entire layers, or part of the whole neural network (in special case). Target labels for the explainer model are (one or more) key channel parameters, such as SNR, Doppler spread, and delay spread. Explainer model will be trained post-hoc. Predictive performance of the explainer model is seen as a proxy for how much information a unit or units of the performer model contains about (one or more) key channel parameters.

The proposed method provides interpretations on both global and local levels. \textit{Global interpretations}---predictive performance of the explainer model for the whole dataset---give insights into which units contain most (or least) information about the key channel parameter(s) on average across the dataset, and are an aggregate of local interpretations. \textit{Local interpretations}---predictive performance of the explainer model for a data instance---offer insights into which units contain most (or least) information about the key channel parameter(s) for individual data instances.

\section{Experiments and Results}
\label{sec:experimentsresults}
We applied the proposed method to interpret how the deep neural network-based receiver model processes the SNR channel parameter at both the layer and convolutional channel levels using the generated dataset. The detailed steps for extracting interpretations are shown in Figure \ref{fig:proposed_method}.

\subsection{Experimental Setup} 
\label{ssec:experimentalsetup}
We generated training and testing data using link-level simulations with the NVIDIA Sionna library \cite{hoydis2023sionna}, covering SNR values from -10dB to 25dB with 192 subcarriers. SNR is defined as the ratio of signal power to noise power, i.e. $\mathrm{SNR} \coloneq \frac{P_\mathrm{signal}}{P_\mathrm{noise}}$. Higher SNR values indicate better signal quality due to a lower noise presence.

The performer model was based on the deep neural network architecture from \cite{honkala2021deeprx}, featuring 5 preactivated ResNet layers, as detailed in Table \ref{table:nn_architecture_parameters}. We trained the performer model following the procedure described in the original paper. The explainer model had a similar architecture but was enhanced with additional fully connected layers. It included a 2D convolutional layer with 64 channels, followed by 4 preactivated ResNet layers and 4 fully connected layers, as outlined in Table \ref{table:nn_architecture_parameters}. We trained the explainer model using PyTorch library \cite{PytorchNeurIPS2019}, with the hyperparameters specified in Table \ref{table:explainer_training_parameters}.

We evaluated the explainer’s predictive performance using Mean Squared Error (MSE) \eqref{mse}. MSE is defined as the average of the squared differences between the true $\mathrm{SNR}_i$ values and the predicted $\widehat{\mathrm{SNR}}_i$ values, and it penalizes larger prediction errors more heavily since differences are squared \cite{murphy2022}.

\begin{equation}\label{mse}
  \mathrm{MSE}(\mathrm{SNR}, \widehat{\mathrm{SNR}}) \coloneq \frac{1}{N}\sum_{i=0}^{N - 1} (\mathrm{SNR}_i - \widehat{\mathrm{SNR}}_i)^2
\end{equation}

As a baseline, we employed the KSG estimator \cite{kraskov2004} with five nearest neighbors to estimate Normalized Mutual Information (NMI) via relative entropy, following Nagel et al. \cite{nagel2024nmi}. To avoid numerical overflow, we applied a logarithmic transformation to the scaling-invariant k-NN radius\footref{additionalresultsnote}. To improve efficiency and address numerical issues, we first applied Principal Component Analysis (PCA) \cite{hotelling1933pca} to retain 95\% of the variance in the activations and further experimented with reducing dimensionality to minimal levels using Uniform Manifold Approximation and Projection (UMAP) \cite{mcinnes2018umap} after PCA reduction.

\begin{table}[!t]
\renewcommand{\arraystretch}{1.3}
\caption{Neural Network Architecture Parameters}
\label{table:nn_architecture_parameters}
\centering
\begin{tabular}{c|c|c}
    \hline
    \bfseries Parameter & \shortstack{\bfseries Performer\\Model} & \shortstack{\bfseries Explainer\\Model}\\
    \hline\hline
    \# of ResNet Layers & 5 & 4 \\
    \hline
    \shortstack{\# of Convolutional Filters\\per ResNet layer} & 64, 64, 32, 32, 32 & 64, 32, 32, 16 \\
    \hline
    Dilations in frequency & 1, 4, 8, 4, 1 & - \\
    \hline
    Dilations in time & 1, 2, 3, 2, 1 & - \\
    \hline
    Fully Connected Layers & - & 1024, 256, 64, 1 \\
    \hline
\end{tabular}
\end{table}

\begin{table}[!t]
\renewcommand{\arraystretch}{1.3}
\caption{Training Parameters for explainer model}
\label{table:explainer_training_parameters}
\centering
\begin{tabular}{c|c}
    \hline
    \bfseries Parameter & \bfseries Explainer Model \\
    \hline\hline
    Optimizer & AdamW \\
    \hline
    Loss Function & MSE \\
    \hline
    Early Stop Tolerance & 20 epochs \\
    \hline
    Batch Size & 128 \\
    \hline
    Learning Rate & $1 \times 10^{-4}$ \\
    \hline
    Weight Decay & $5 \times 10^{-4}$ \\
    \hline
\end{tabular}
\end{table}

\subsection{Results}
\label{ssec:results}

\begin{figure*}[t]
\centering
\includegraphics[width=\textwidth]{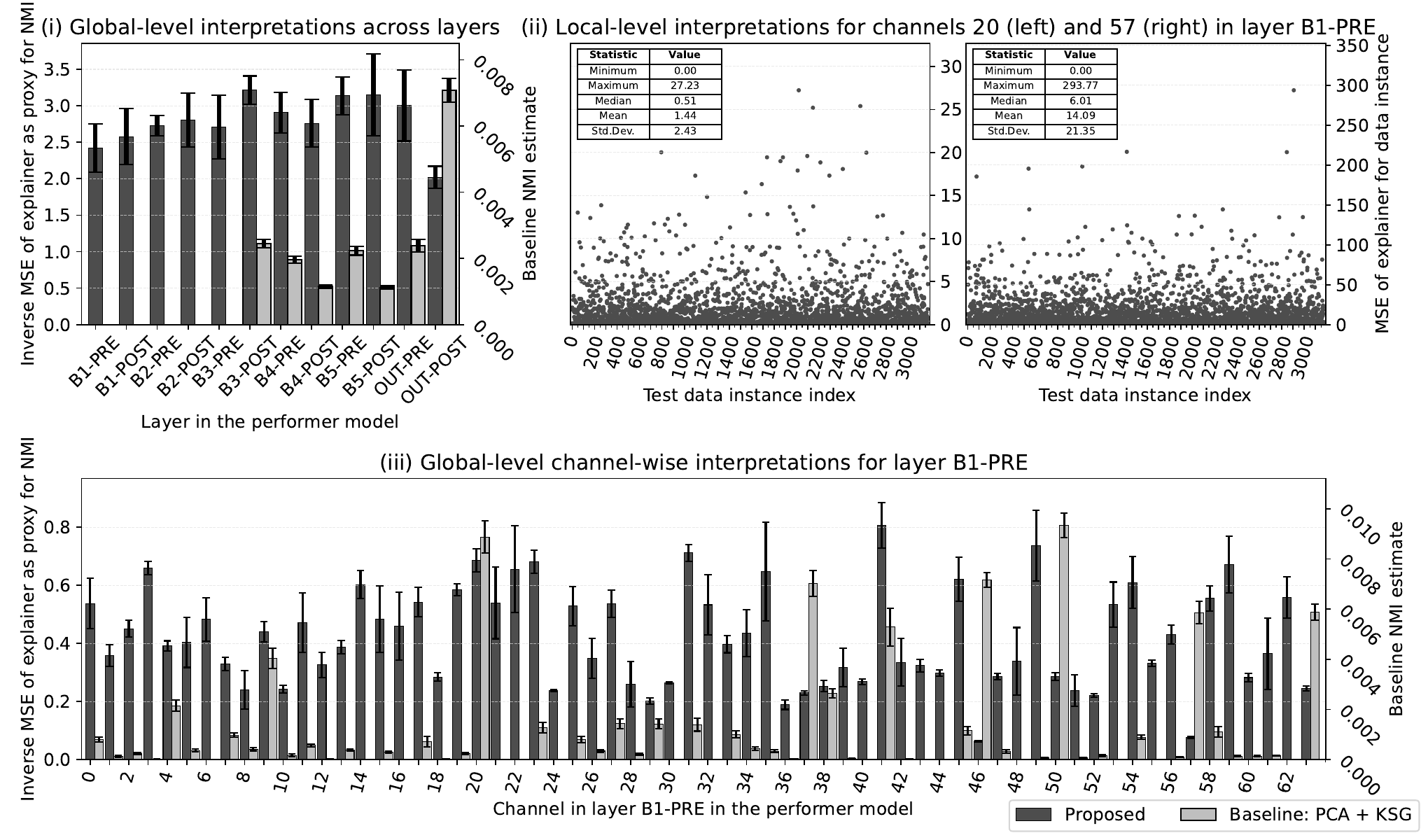}
\caption{Interpretations of selected ResNet layers and channels for the the deep neural network-based receiver model as performer model. Subplots (i) and (iii) show the means of the global interpretations across layers and channel-specific insights for layer \textit{B1-PRE}, respectively, based on ten different data folds with standard deviations indicated by error bars. Subplot (ii) show local, instance-specific interpretations for channels 20 and 57 in layer \textit{B1-PRE}. Activations in pre-activation ResNet blocks are denoted as follows: \textit{B1-PRE} refers to activations after the first ReLU in the first ResNet block, while \textit{B1-POST} refers to activations after the second ReLU in the same block. This naming convention continues similarly for subsequent blocks.}
\label{fig:experimental_results}
\end{figure*}

According to our experiments, the proposed method is not overly data-dependent. Thus, the observed phenomenon---explainer model’s predictive
performance---is dependent on the model rather than the data, which is supported by the reasonable standard deviations observed in the k-fold cross-validation shown in Figures \ref{fig:experimental_results}(i) and \ref{fig:experimental_results}(iii). Additionally, the training process with different seeds appears stable\footnote{\label{additionalresultsnote}Additional material can be found in the Appendix, see Section~\ref{sec:appendix}.}.

Considering that a higher inverse MSE of the explainer model indicates a more informative unit in the performer model, we observe that, on average, different layers contain diverse amount of information on SNR, with middle layers being more informative, as illustrated in Figure \ref{fig:experimental_results}(i). This aligns with the established understanding that intermediate neural network layers, when evaluated by external criteria, are more informative than those in the input or output \cite{goodfellow2016deep}. Furthermore, different channels exhibit an even greater diversity in the amount of information they provide on SNR, especially channels 46 and 57 in layer \textit{B1-PRE} seems to contain much less information on SNR than other channels in the layer, as shown in Figure \ref{fig:experimental_results}(iii).

To examine the less informative channel 57 in layer \textit{B1-PRE}, we visualized it alongside channel 20, which we found to contain more information on SNR. Figure \ref{fig:experimental_results}(ii) shows that channel 57 exhibits about 10 times higher intra-instance maximum MSE, mean, and deviation compared to channel 20. Our further analysis\footref{additionalresultsnote} revealed highly skewed, right-tailed distributions of the local level interpretations---both across different layers, and across different channels---indicating significant variability and potential outliers, with a few instances disproportionately impacting the global level interpretations.

The baseline method produced consistently low or zero NMI estimates, suggesting inaccurate estimation, especially compared to the success of our proposed method in predicting SNRs. The high-dimensionality, even after PCA, likely overwhelmed the KSG estimator, which---despite its ability to capture complex relationships \cite{kraskov2004}---is vulnerable to the curse of dimensionality \cite{bellman1961}. Reducing dimensionality to single digits using UMAP after PCA made estimates\footref{additionalresultsnote} highly dependent on final dimensionality, with dimensions of ten or higher causing numerical issues, including negative entropies, as known to happen with too few samples or outliers \cite{GitHubNorMIissue6}.

\section{Discussion}
\label{sec:discussion}
The experimental results suggest that channels 46 and 57 in layer \textit{B1-PRE} are non-beneficial for SNR processing. Therefore, it may be possible to remove these channels without negatively affecting performance, leading to a more compact performer model. However, since a few instances disproportionately impact interpretations, focusing efforts on these key instances---such as further analysis or development---may also be beneficial.

The proposed method shows low data dependence, suggesting strong generalizability. As a result, the model’s performance is expected to remain consistent across different datasets and channel conditions. Future work should explore and confirm this generalizability.

The robustness of neural network interpretations is closely tied to the robustness of the network itself \cite{ghorbani2019fragility}. Therefore, the performance and fragility of both the explainer and performer models are interconnected. Future work should focus on leveraging information about the predictive performance and fragility of the performer model when interpreting results from the explainer model, while also enhancing the robustness and generalizability of both models to address potential fragility issues.

\section{Conclusion}
\label{sec:conclusion}
The proposed method provides interpretations for neural networks at both global and local levels, with global explanations being aggregations of local explanations. This dual-level approach allows users to understand overall trends or focus on specific cases, enhancing their ability to identify and address issues, thereby improving overall model performance.

Experimental results on a deep neural network-based receiver model and SNR channel parameter demonstrate that the proposed method is not overly data-dependent and offers robust estimation in high-dimensional settings where baseline methods struggle. These results highlight the method's effectiveness in providing insights into the model's internal processing under varying SNRs, which can be leveraged to improve and troubleshoot the model.

Future work will explore the proposed method’s applicability to other channel parameters, test its performance on real-world datasets, and further investigate how to incorporate information about the predictive performance of the performer model and the fragility of both the performer and explainer models into the interpretation process.

\clearpage
\section{Appendix}
\label{sec:appendix}

\subsection{Improve Numerical Stability in NMI Estimation}
\label{ssec:improvenumericalstability}
To address numerical overflow issues during the estimation of Normalized Mutual Information (NMI) in high-dimensional spaces, we propose applying a logarithmic transformation to the calculation of the scaling-invariant k-nn radius. This transformation enhances the stability of the radius computation without compromising precision. Our approach has been validated through both theoretical and empirical analysis, with detailed results presented in our manuscript, “Improving Numerical Stability of Normalized Mutual Information Estimator on High Dimensions,” available as a preprint on arXiv at \href{https://arxiv.org/abs/2410.07642}{arXiv:2410.07642}.

\subsection{Stability of Training with Different Seeds}
\label{ssec:stabilityoftraining}
\begin{figure*}[t]
\centering
\includegraphics[width=\textwidth]{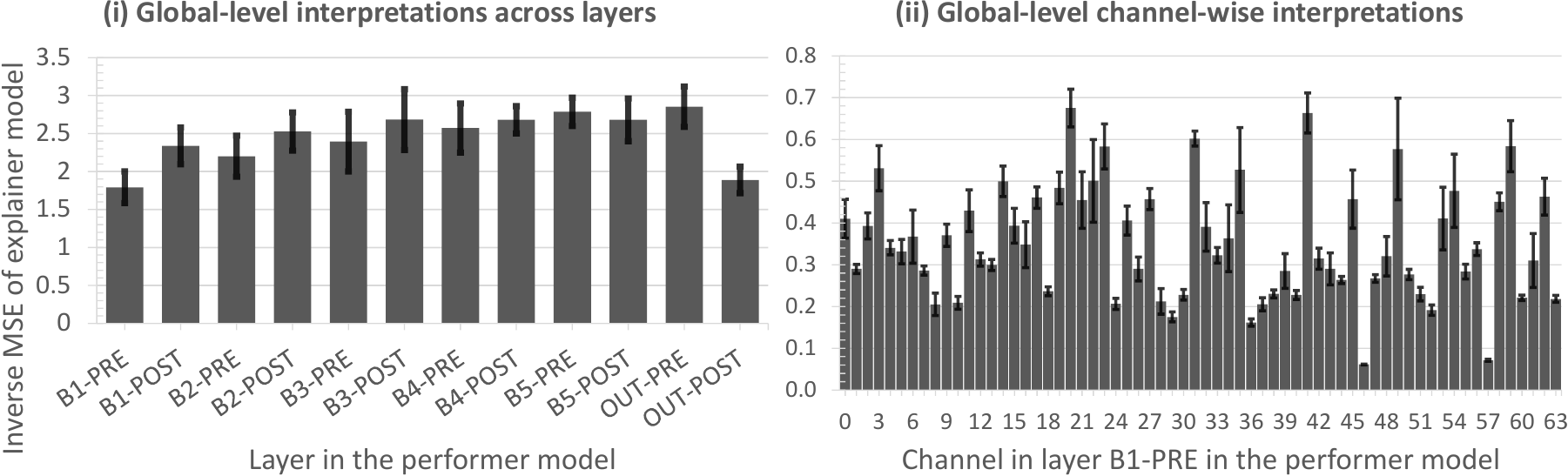}
\caption{Interpretations of selected ResNet layers and channels for the the deep neural network-based receiver model as performer model. Subplots (i) and (ii) show the means of the global interpretations across layers and channel-specific insights for layer \textit{B1-PRE}, respectively, \textbf{based on ten different random seeds} with standard deviations indicated by error bars. The naming convention for layers follows that of Figure~\ref{fig:experimental_results}.}
\label{fig:experimental_results_seeds}
\end{figure*}

The training process of the performer model was repeated using different random seeds to evaluate its stability. Across these experiments, the method demonstrated stable performance, as indicated by the reasonable standard deviations observed in the layerwise results shown in Figure~\ref{fig:experimental_results_seeds}(i) and the channelwise results in Figure~\ref{fig:experimental_results_seeds}(ii). These results highlight the consistency of the model's behavior when trained with the same data fold but initialized with ten different random seeds, supporting the robustness of the training process.

\subsection{Further Analysis of Local-Level Interpretations}
\label{ssec:analysisofskewed}
We extracted local-level interpretations from the proposed method, which revealed highly skewed, right-tailed distributions---both across different layers and channels. These distributions indicate significant variability and the presence of potential outliers, with a small number of data instances disproportionately influencing the global-level interpretations. 

\begin{figure*}[t]
\centering
\includegraphics[width=\textwidth]{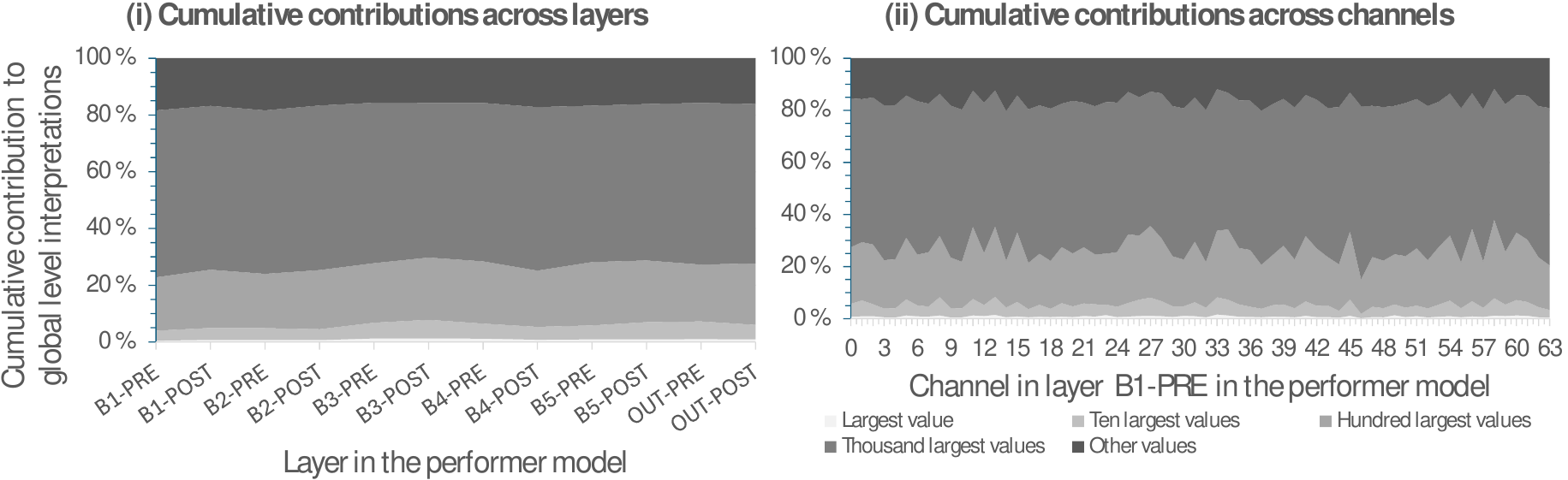}
\caption{Cumulative contributions of test data instances to the global-level interpretations for the deep neural network-based receiver model as the performer model. Subplot (i) shows the cumulative contributions across layers, while subplot (ii) presents the cumulative contributions across channels for layer \textit{B1-PRE}. The naming convention for layers follows that of Figure~\ref{fig:experimental_results}.
Contributions are displayed for the largest value, the ten largest values, the one hundred largest values, the one thousand largest values, and the remaining data instances. The contribution of each data instance is calculated as its individual value divided by the total sum of all data instances, reflecting its relative importance to the global-level interpretations.}
\label{fig:cumulative_contributions}
\end{figure*}

Figure~\ref{fig:cumulative_contributions}(i) illustrates the cumulative contributions across layers, while Figure~\ref{fig:cumulative_contributions}(ii) shows the cumulative contributions across channels for layer \textit{B1-PRE}. The contribution of each data instance was calculated as its individual value divided by the total sum of all data instances, reflecting its relative importance to the global-level interpretations. Since global-level interpretations are aggregates of local-level interpretations, this relative importance also constitutes the actual contribution of each instance to the global-level result. The cumulative contribution of the $n$ largest values is then the sum of the contributions of the $n$ highest-contributing data instances.

For example, Figure~\ref{fig:cumulative_contributions}(ii) shows that no single outlier dominates, with the largest individual contribution typically accounting for around 1\%. However, the top 1,000 values contribute over 80\%, indicating that a relatively small portion (about 31\%) of the data accounts for the majority of the interpretation variability---characteristic of long-tailed distributions. This analysis reveals two key findings: (1) local-level interpretations exhibit significant variability and potential outliers, and (2) a small number of instances have a disproportionately large impact on the global-level interpretations.

\subsection{Impact of UMAP Dimensionality Reduction}
\label{ssec:impactofumap}
The baseline method, which used the Kraskov-Stögbauer-Grassberger (KSG) estimator for Normalized Mutual Information (NMI) estimation, consistently produced low or zero NMI estimates, even after applying Principal Component Analysis (PCA) to reduce dimensionality. This suggests inaccurate estimation, as the baseline method failed to capture meaningful mutual information between variables. We suspect that the KSG estimator struggled to capture certain non-linear dependencies present in the data. The high-dimensional nature of the data, even after PCA reduction, likely overwhelmed the estimator, leading to poor NMI estimates---particularly in contrast to the success of our proposed method in predicting SNRs, as discussed in Section~\ref{ssec:results}.

To further investigate, we applied a two-step dimensionality reduction, first using PCA to retain 95\% of the variance, followed by Uniform Manifold Approximation and Projection (UMAP) to further reduce dimensionality. Since the baseline method had previously failed with activations from the \textit{OUT-POST} layer, we focused on these activations, experimenting with dimensionality reductions from one to eleven dimensions.

\begin{figure*}[t]
\centering
\includegraphics[width=\columnwidth]{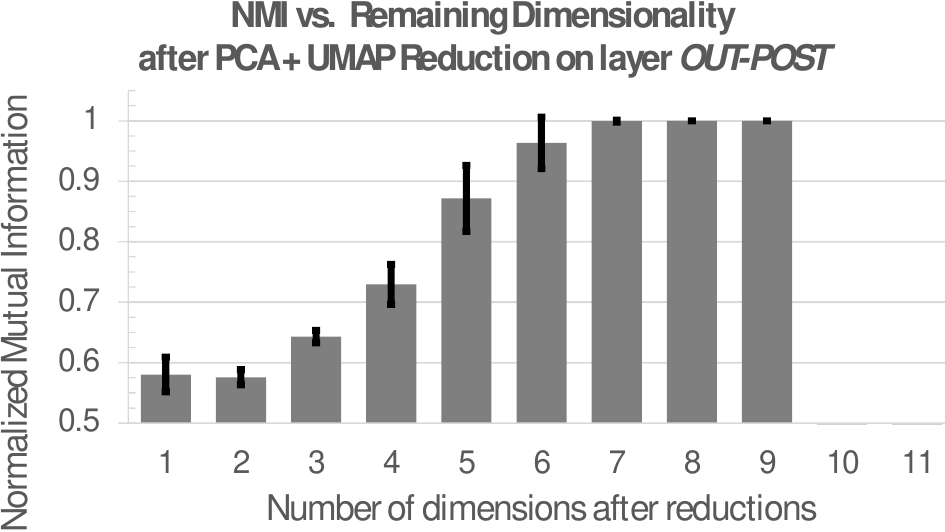}
\caption{Normalized Mutual Information (NMI) estimates based on ten different data folds with standard deviations indicated by error bars, as a function of the number of remaining dimensions on the \textit{OUT-POST} layer after a two-step reduction process. The process first reduces dimensionality using PCA, retaining 95\% of the variance, and then applies UMAP to further reduce dimensions from one to eleven. The method fails completely due to negative entropies for ten or more dimensions. The naming convention for layers follows that of Figure~\ref{fig:experimental_results}.}
\label{fig:nmi_vs_dimensionality}
\end{figure*}

\begin{figure*}[t]
\centering
\includegraphics[width=\textwidth]{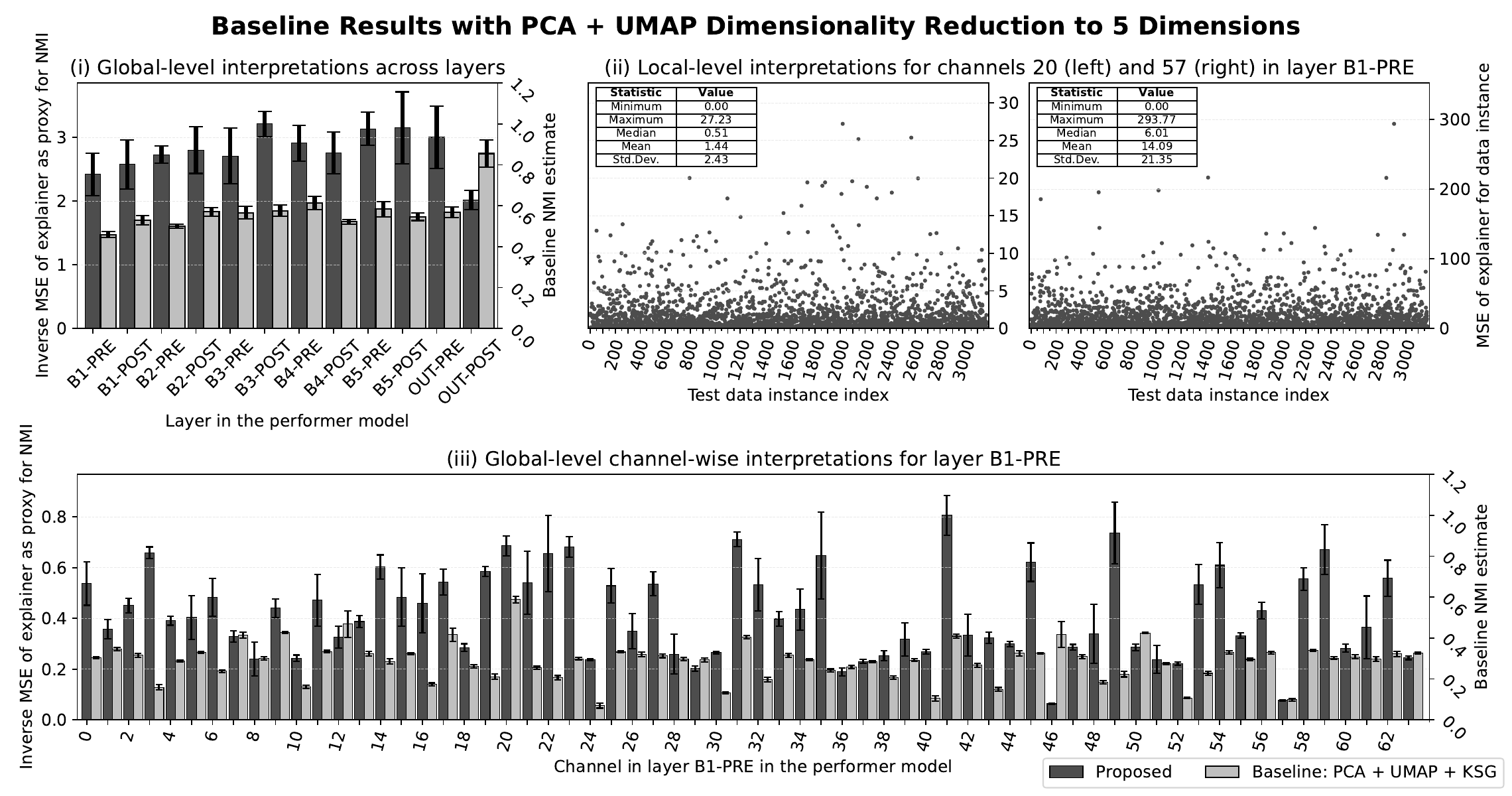}
\caption{Baseline results based on ten different data folds, with standard deviations indicated by error bars, after applying PCA (retaining 95\% of variance) followed by UMAP to reduce dimensionality to five dimensions. Comparing to Figure~\ref{fig:experimental_results_umap10}, it is clear that the estimates are highly dependent on the final dimensionality. The naming convention for layers follows that of Figure~\ref{fig:experimental_results}.}
\label{fig:experimental_results_umap5}
\end{figure*}

\begin{figure*}[t]
\centering
\includegraphics[width=\textwidth]{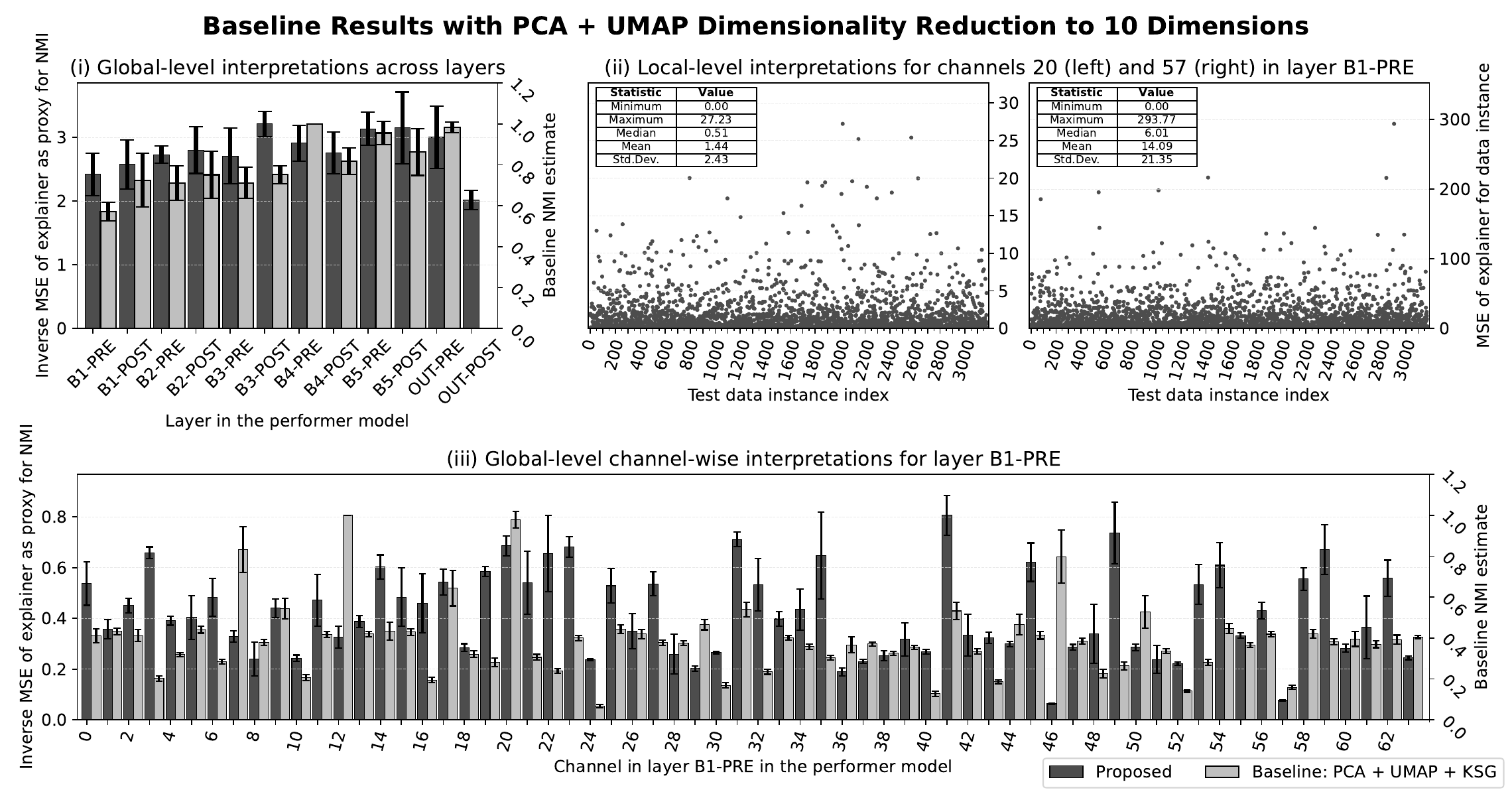}
\caption{Baseline results based on ten different data folds, with standard deviations indicated by error bars, after applying PCA (retaining 95\% of variance) followed by UMAP to reduce dimensionality to ten dimensions. Comparing to Figure~\ref{fig:experimental_results_umap5}, it is clear that the estimates are highly dependent on the final dimensionality. The method fails completely on the \textit{OUT-POST} layer, as all runs resulted in failure due to negative entropies. The naming convention for layers follows that of Figure~\ref{fig:experimental_results}.}
\label{fig:experimental_results_umap10}
\end{figure*}

Figure~\ref{fig:nmi_vs_dimensionality} illustrates the NMI estimates as a function of dimensionality. We observed a steady increase in NMI with higher dimensions; however, beyond ten dimensions, the method began to fail due to negative entropies---an issue often caused by insufficient data or the presence of outliers \cite{GitHubNorMIissue6}. Furthermore, the baseline method was highly sensitive to the dimensionality after reduction, as demonstrated in Figures~\ref{fig:experimental_results_umap5} and \ref{fig:experimental_results_umap10}, which show results for reductions to five and ten dimensions, respectively.

Using the KSG estimator for MI estimation is a classical approach for estimating mutual information. Similarly, PCA and UMAP, whether applied individually or together, are well-established classical methods for the dimensionality reduction. Therefore, applying PCA+UMAP for dimensionality reduction followed by the KSG estimator for MI estimation represents a classical approach as a whole. However, as our results demonstrate, this classical approach may not be effective in certain contexts, particularly when dealing with high-dimensional data (and potentially when non-linear dependencies exist between variables).

Our conclusion is that the classical approach of using PCA+UMAP for dimensionality reduction combined with the KSG estimator for MI estimation is unreliable and not robust in this context. This is likely due to its sensitivity to parameter selection, particularly the final number of dimensions. While PCA+UMAP and KSG are often effective in many situations, our results suggest that this classical trick does not resolve the underlying issues here and may even exacerbate them. Although the results regarding non-linear dependencies were inconclusive, we found that the baseline method is highly dependent on the dimensionality after reduction and is not suitable for estimating NMI between activations and the SNR channel parameter.

\clearpage

\clearpage
\bibliographystyle{IEEEtran}
\bibliography{IEEEabrv,refs}

\end{document}